\documentclass[11pt,a4paper]{article}
\usepackage[]{emnlp2021}
\usepackage{times}
\usepackage{latexsym}

\usepackage{microtype}

\usepackage{booktabs}
\usepackage{multirow, multicol}
\usepackage{tabularx}
\usepackage[inline]{enumitem}
\usepackage[export]{adjustbox}
\usepackage{microtype}
\usepackage{rotating}

\usepackage{mathtools}
\usepackage{subcaption} 
\usepackage[skip=5pt]{caption}
\usepackage{commath}
\usepackage{enumitem,color, amssymb}
\usepackage{pifont}
\usepackage{amsmath,amsfonts,amsthm}

\usepackage{enumitem}
\usepackage[super]{nth}
\usepackage{pgf}
\usepackage{graphicx}
\graphicspath{{./images/}} 

\usepackage{appendix}
\usepackage{algorithm}
\usepackage[noend]{algpseudocode}
\usepackage[subtle]{savetrees}
\usepackage{lipsum}

\usepackage{tikz}
\usepackage{pgfplots}
\setlist[itemize]{leftmargin=*}
\usepackage[disable]{todonotes}
\newcommand{\howmanylanguages}{20}%
\newlist{romanlist}{enumerate*}{1}
\setlist[romanlist]{label=(\roman*)}

\newcommand{\adi}[2][]{\todo[author=adi,fancyline,size=\footnotesize,color=green,#1]{#2}}
\newcommand{\Adi}[2][]{\adi[inline,#1]{#2}\noindent}

\title{Investigating Failures of Automatic Translation\\ in the Case of Unambiguous Gender}

\author{Adithya Renduchintala \\
  Facebook AI \\
  \texttt{adirendu@fb.com} \\\And
  Adina Williams \\
  Facebook AI Research \\
  \texttt{adinawilliams@fb.com} \\}

\date{}

\begin{document}
\maketitle
\begin{abstract}
Transformer based models are the modern work horses for neural machine translation (NMT), reaching state of the art across several benchmarks.
Despite their impressive accuracy, we observe a systemic and rudimentary class of errors made by transformer based models with regards to translating from a language that doesn't mark gender on nouns into others that do.
We find that even when the surrounding context provides unambiguous evidence of the appropriate grammatical gender marking, no transformer based model we tested was able to accurately gender occupation nouns systematically.
We release an evaluation scheme and dataset for measuring the ability of transformer based NMT models to translate gender morphology correctly in unambiguous contexts across syntactically diverse sentences.
Our dataset translates from an English source into $20$ languages from several different language families. 
With the availability of this dataset, our hope is that the NMT community can iterate on solutions for this class of especially egregious errors.
\end{abstract}
\section{Introduction}

NMT models have come a long way since their widespread adoption. Modern Transformer based \citep{vaswani-etal-2017-attention} NMT architectures can be trained on vast amounts of data, and are constantly attaining higher BLEU scores on standard benchmarks~\cite{barrault-etal-2020-findings}. Despite this impressive performance, we observed that state-of-the-art Transformer-based MT systems are largely unable to make basic deductions regarding how to correctly inflect nouns with grammatical gender, \emph{even} when there is ample contextual evidence. For example, we observe that they struggle inflect occupation nouns like ``doctor'' with the correct gender when translating sentences like ``my mother is a funny doctor'', despite there being no ambiguity that ``doctor'' ought to be marked with feminine. This suggests that our current training paradigm (or the architectures themselves), do not force models to pay sufficient attention to very basic linguistic properties of a source sentence during inference. When an NMT model makes mistakes like these, it can degrade the trust a user places on translation quality, or also reinforce representational harms in the form of stereotypes \citep{stanovsky-etal-2019-evaluating}. 

To more systematically explore translation failures for gender in unambiguous sentences, we have created a benchmark dataset that clearly surfaces these kinds of errors hoping that the wider NMT community can devise better, targeted mitigation strategies. Our benchmark contains specially constructed English source sentences which unambiguously belie the gender of a person referred to with a noun that can be inflected for multiple grammatical genders. In our setting, we use sentences containing occupation nouns from English (which do not bear grammatical gender marking in the source language) and translate them into languages for which occupation nouns must bear grammatical gender.  

We craft unambiguous source sentences by manipulating the context: In our dataset,  an unambiguously gendered word---i.e., a noun (such as \textit{father}) or pronoun (such as \textit{herself})---obligatorily corefers with the occupation noun making it clear what the gender on the occupation noun must be. Consider ``My nurse is a good \textit{father}'', ``I am a nurse who can inspire \textit{herself}''. Although all our occupations nouns are technically underspecified (i.e., able to refer to any gender in the source language), we also vary whether the occupation noun is stereotypically more likely to refer to a man or woman (e.g., \textit{janitor} vs. \textit{nurse}). Finally, we vary whether the triggering word appears before or after the occupation noun, to see if this affects performance: compare ``That janitor arrives early for \textit{her} shift'' to  ``That \textit{her} own child laughed surprised the janitor''. 

To show the utility of our benchmark, we evaluate the accuracy of the gender inflection in the target language for several state-of-the art Transfomer based NMT systems. 
Previous work focuses primarily on ambiguous cases where the gender of an occupation noun is genuinely under-determined \citep{stanovsky-etal-2019-evaluating}, since this allows for querying the underlying (often stereotypical) ``assumptions'' of the translation model. We argue that our unambiguous task is an even clearer explication of the failures of NMT models, because in this set up, morphological gender mistakes are not forgivable. 

Because we expect existing MT systems to struggle on our unambiguous task, we also devised a somewhat simpler setting. It is well known that translation systems perform better when provided with more context (i.e., longer sentences; \citealt{tiedemann-scherrer-2017-neural,miculicich-etal-2018-document}), so in our simpler setting, we augment our sentences with statistically gender-indicative adjectives (such as \textit{pretty} and \textit{handsome}, the former being used more often in practice to modify nouns referring to women and the latter, to men) and verbs (which take arguments that are statistically more likely to be men or women in large corpora). With these sentences, we can determine how much the determination of the gender on gender-indeterminate nouns is affected by statistically gender-indicative contexts. We expect the incidence of correct inflection to rise in cases when a stereotypical contextual cue is also provided.

Our contributions are as follows: We offer a new unambiguous benchmark to measure MT models' ability to mark gender appropriately in $20$ languages from English source. We find that all tested Transformer-based NMT models reach fairly low accuracy---at best approximately 70\% accuracy (Portuguese and German) and at worst below 50\% (Urdu)---and do markedly better when the word that makes the target gender explicit (e.g., \textit{her}, \textit{brother}) refers to a man as opposed to a woman. Moreover, we find that accuracy is higher on examples for which the (statically more frequent) gender of the occupation matches the gender of the unambiguous triggering word, compared to examples for which they don't match. 

\section{Methods}

\begin{table}[ht]
\centering
\begin{adjustbox}{max width=\columnwidth}
\begin{tabular}{@{}lll@{}}
\toprule
\multicolumn{2}{l}{\textbf{Source/Target}}               & \textbf{Label}                \\ \midrule
Src: & My sister is a carpenter$_{4}$ .               & \multirow{2}{*}{Correct}      \\
Tgt: & Mi hermana es carpenteria(f)$_{4}$ .           &                               \\
Src: & That nurse$_{1}$ is a funny man .              & \multirow{2}{*}{Wrong}        \\
Tgt: & Esa enfermera(f)$_{1}$ es un tipo gracioso .   &                               \\
Src: & The engineer$_{1}$ is her emotional mother .   & \multirow{2}{*}{Inconclusive} \\
Tgt: & La ingeniería(?)$_{1}$ es su madre emocional . &                               \\ \bottomrule
\end{tabular}
\end{adjustbox}
\caption{Examples of source-translation pairs. The gender-tags are shown in parenthesis and word-alignments indicated with subscript.}
\label{tab:exampletranslations}
\end{table}

Our method crucially relies upon on linguistic theory to craft unambiguous examples. In most attempts to measure gender bias in NMT, there has been no ground-truth ``correct translation''---model ``preferences'' \citep{stanovsky-etal-2019-evaluating, prates2019assessing} are reflected by the percentage of examples for which the MT system chooses the gender-stereotypical pronoun as opposed to the anti-gender-stereotypical. However, since both translations are practically possible in reality (for example, janitors come in all genders), we feel this setting might be overly optimistic about the capabilities of current models. 

Our set up has two main components: we have a ``trigger'' (i.e., a noun or pronoun in the source sentence that unambiguously refers to a person with a particular known gender\footnote{Gender identity is not strictly binary, and even for our strongly gendered triggers, there still could be rare edge-cases: consider, for example, a Halloween party where your friend Sam, who identifies as a man, dresses up as his own grandmother. Someone can reasonably refer to Sam during the party as a ``grandmother"  or choose either ``she'' or ``he''; See also \citep{ackerman2019syntactic}.}), and we have an occupation noun which bears no gender-marking in the source language and can be inflected with various genders in the target language. We call the former class ``triggers'' because they are the unambiguous signal which \emph{triggers} a particular grammatical gender marking for the occupation noun. Triggers comprise all ``standard'' American English pronouns and explicitly gendered kinship terms, which were chosen because they are very common concepts cross-linguistically and are (in nearly all cases, see fn.~1) interpreted as gender-unambiguous. Occupation nouns were drawn from the U.S. Bureau of Labor Statistics\footnote{\url{http://www.bls.gov/cps/cpsaat11.htm}}, following \newcite{caliskan2017,rudinger-etal-2017-social, zhao-etal-2018-gender, prates2019assessing}, and are  statistically more likely to be performed by either women or by men respectively. We ensure that there is an equal number of triggers, occupation words, making our benchmark gender-balanced for binary gender. For a list, see \autoref{tab:triggeroccupation}, and \autoref{tab:contextcues} in the Appendix.

Crucially, we measure performance based on the inflection of the occupation noun, which depends on the syntactic structure of the sentence. To ensure that we have unambiguous sentences, we constructed a short English phrase structure grammar comprising 82 commands to construct our corpus. Although previous datasets for measuring gender failures in translation have had a handful unambiguous examples \citep{stanovsky-etal-2019-evaluating}, our dataset is unique in having \emph{only} unambiguous examples (see also \citealt{gonzalez-etal-2020-type}). We also make use of Binding Theory \citep{chomsky1980binding, chomsky1981lectures, buring2005binding} to ensure that (i) all of our pronoun triggers (both pronominals like ``she'' and anaphors like ``herself'') are strictly coreferring with the occupations and (ii) that no other interpretations are possible.\footnote{Consider the sentence ``Carlotta's dog accompanies her to kindergarden'' \citep[p.5]{buring2005binding}. In this sentence, we can interpret this sentence as meaning that the dog accompanies Carlotta to kindergarden, or that the dog accompanies some other woman or girl to kindergarden---to strengthen this reading you can append to the front of the sentence the clause something like ``whenever Mary's parents have to go to work early, Carlotta's dog accompanies her to kindergarden''. In this way, ``her'' can refer to either Carlotta or to Mary. We have avoided such ambiguity in our dataset.} Moreover, having a grammar is useful, since it allows for an increased diversity of source sentences and better control over the context. 

Since we anticipated poor performance on the task, we also devised an easier scenario, where we provide additionally contextual cues to the gender of the relevant entity. In this work, we explore two types of contextual cues, adjectives and verbs. Our list of adjectives is the union of stereotyped traits from several works in the social psychology literature on traits as gender stereotyping \citep{bem1981bem,prentice2002women,haines2016times,eagly2020gender}, where they were normed in the context of English. Verbs were automatically discovered from Wikipedia using dependency parses to find verbs that take women or men preferentially highly as subjects or direct objects \citep{hoyle-etal-2019-unsupervised}.

Finally, given that we had already written a toy grammar for English, we also craft our grammar to enable the exploration of a couple subsidiary questions about the nature of anaphoric relations: for example, does accuracy depend on whether the occupation precedes or follows the trigger? Moreover, when we include a contextual cue that is predictive of the gender required by the trigger (e.g., \textit{handsome} for \textit{brother}), does accuracy change when we attach it to the occupation (e.g., \textit{that handsome nurse is my brother}) instead of to the trigger (\textit{that nurse is my handsome brother})? And finally, to what extent do these different syntactic factors interact with each other or vary across languages? 

\begin{table}[t]
\small
\begin{tabular}{p{4em}p{8em}p{8em}}
\toprule
Type &  F & M   \\
\midrule
 Trigger & she, her, hers, herself, sister , mother, aunt, grandmother, daughter, niece, wife , girlfriend & he, him, his, himself, brother, father, uncle, grandfather, son, niece, husband, boyfriend\\
 Occupation & editor, accountant, auditor, attendant, assistant,  designer, writer, baker, clerk, cashier, counselor, librarian, teacher, cleaner, housekeeper, nurse, receptionist, hairdresser, secretary & engineer, physician, plumber, carpenter, laborer, driver, sheriff, mover, developer, farmer, guard, chief, janitor, lawyer, CEO, analyst, manager, supervisor, salesperson \\
\bottomrule
\end{tabular}
    \caption{Gendered words from our dataset. Accuracy is measured on the occupation word, and the Trigger(s) provide unambiguous information about the gender identity of the person being referred to in the sentence. Establishing co-reference between the two is obligatory, based on the syntactic structures included in the dataset.}
    \label{tab:triggeroccupation}
\end{table}

\subsection{Models}
We evaluate gendered translation of two pretrained open-source models,
\begin{romanlist}
    \item \emph{OPUS-MT} is a collection of 1000+ bilingual and multilingual (for certain translation directions) models~\cite{tiedemann-thottingal-2020-opus}. The architecture of each model was based on a standard transformer~\cite{vaswani-etal-2017-attention} setup with 6 self-attentive layers in both, the encoder and decoder network with 8 attention heads in each layer.
    \item \emph{M2M-100} is a large multilingual model which supports ``many-to-many'' translation directions~\cite{fan2020beyond}. M2M-100 pretrained models are available in three sizes (418 Million parameters, 1.2 Billion parameters and 15 Billion parameters). We employ the small 
    sized models for our experiments which are based on the transformer architecture with 12 encoder and decoder layers and 16 attention heads. 
\end{romanlist}
\Adi{double-check the number of heads}
\Adi{removed mBART-50 because we have no numbers from it...}

\subsection{Evaluation}
Using our grammar, we generate English source sentences and translate these into supported target languages. 
To ascertain whether the translation applied the correct morphological marker on the target-side occupation noun, we design a ``reference-free'' evaluation scheme.
Following \newcite{stanovsky-etal-2019-evaluating}, we extract token-alignments between the source occupation noun token and its translation in the target side. 
We also extract morphological features for every token in the target sequence, using a morphological tagger.
Thus, we can ascertain the gender associated with the translated occupation noun (as judged by the morphological tagger) and measure the NMT models' accuracy concerning gender translation.
We use \newcite{dou2021word} for word-alignment and \newcite{qi-etal-2020-stanza} as our morphological tagger.
Note that our evaluation scheme only checks if the appropriate gender marking is applied on the occupation noun and does not check if the occupation noun itself has been translated correctly. Thus, we do not prescribe our evaluation scheme as a replacement for traditional MT evaluation using BLEU or chrF++ scores~\cite{papineni-etal-2002-bleu,popovic-2015-chrf}.

Under our evaluation scheme, there are three possible evaluation outcomes for each sentence. We deem the output 
\begin{romanlist}
    \item \emph{correct} if the gender of the target-side occupation noun is the expected gender (based on the source-side trigger gender).
    \item \emph{wrong} if the gender of the target-side occupation is \emph{explicitly} the wrong gender, and 
    \item \emph{inconclusive} if we are unable to make a gender-determination of the target-side occupation noun.
\end{romanlist}
A translation can result be inconclusive if there are errors in the translation, word-alignments or morphological tagger. In most, cases we find translation errors as the root cause of the inconclusive result. Note: if errors predominate more for one gender, this can also be taken as evidence of an imbalance that needs rectification. 

\Adi{todo, this analysis has not been done}
\captionsetup[]{font=small,skip=0pt}
\usetikzlibrary{patterns}
\begin{figure*}
\centering
  \begin{subfigure}{0.49\textwidth}
\begin{tikzpicture}
    \pgfplotsset{%
        width=\textwidth,
        height=1.5in,
    }
    \tikzstyle{every node}=[font=\small]
	\begin{axis}[ybar stacked,bar width=4,
	symbolic x coords={\textbf{avg},de,pt,pl,cs,ru,lt,lv,uk,hr,fr,el,es,ro,hi,he,it,ur,ca,be,sr},
	axis x line*=bottom,
	axis y line*=left,
    xtick=data,
    ytick distance=0.2,
    ymin=0,
	nodes near coords align={vertical},
    x tick label style={rotate=90,anchor=east},
    ]
\addplot +[fill=green!40,draw=green!100] coordinates { (\textbf{avg},0.84) (de,0.97) (pt,0.95) (pl,0.95) (cs,0.95) (ru,0.95) (lt,0.94) (lv,0.93) (uk,0.92) (hr,0.91) (fr,0.91) (el,0.91) (es,0.84) (ro,0.83) (hi,0.81) (he,0.78) (it,0.77) (ur,0.76) (ca,0.75) (be,0.66) (sr,0.34) };

\addplot +[fill=yellow!40,draw=yellow!100] coordinates { (\textbf{avg},0.10) (de,0.02) (pt,0.03) (pl,0.03) (cs,0.03) (ru,0.03) (lt,0.02) (lv,0.02) (uk,0.04) (hr,0.05) (fr,0.07) (el,0.08) (es,0.15) (ro,0.12) (hi,0.11) (he,0.19) (it,0.21) (ur,0.19) (ca,0.24) (be,0.19) (sr,0.09) };

\addplot +[fill=red!40,draw=red!100] coordinates { (\textbf{avg},0.06) (de,0.02) (pt,0.02) (pl,0.02) (cs,0.02) (ru,0.02) (lt,0.03) (lv,0.05) (uk,0.04) (hr,0.04) (fr,0.02) (el,0.02) (es,0.01) (ro,0.05) (hi,0.08) (he,0.03) (it,0.02) (ur,0.05) (ca,0.01) (be,0.15) (sr,0.58) };

	\end{axis}
\end{tikzpicture}
\caption{M-trigger, M-occupation}
    \label{fig:1}
  \end{subfigure}
  \begin{subfigure}{0.49\textwidth}
\begin{tikzpicture}
    \pgfplotsset{%
        width=\columnwidth,
        height=1.5in,
    }
    \tikzstyle{every node}=[font=\small]
	\begin{axis}[ybar stacked,bar width=4,
	symbolic x coords={\textbf{avg},de,pt,pl,cs,ru,lt,lv,uk,hr,fr,el,es,ro,hi,he,it,ur,ca,be,sr},
	axis x line*=bottom,
	axis y line*=left,
    xtick=data,
    ytick distance=0.2,
    ymin=0,
	nodes near coords align={vertical},
    x tick label style={rotate=90,anchor=east},
    ]
\addplot +[fill=green!40,draw=green!100] coordinates { (\textbf{avg},0.52) (de,0.76) (pt,0.72) (pl,0.59) (cs,0.59) (ru,0.43) (lt,0.51) (lv,0.45) (uk,0.49) (hr,0.61) (fr,0.57) (el,0.41) (es,0.58) (ro,0.43) (hi,0.25) (he,0.54) (it,0.49) (ur,0.14) (ca,0.53) (be,0.50) (sr,0.87) };

\addplot +[fill=yellow!40,draw=yellow!100] coordinates { (\textbf{avg},0.07) (de,0.01) (pt,0.02) (pl,0.02) (cs,0.02) (ru,0.01) (lt,0.01) (lv,0.02) (uk,0.02) (hr,0.03) (fr,0.06) (el,0.07) (es,0.20) (ro,0.04) (hi,0.08) (he,0.07) (it,0.23) (ur,0.19) (ca,0.18) (be,0.14) (sr,0.03) };

\addplot +[fill=red!40,draw=red!100] coordinates { (\textbf{avg},0.40) (de,0.23) (pt,0.26) (pl,0.39) (cs,0.39) (ru,0.56) (lt,0.48) (lv,0.52) (uk,0.50) (hr,0.36) (fr,0.37) (el,0.53) (es,0.22) (ro,0.52) (hi,0.67) (he,0.39) (it,0.28) (ur,0.67) (ca,0.29) (be,0.36) (sr,0.10) };
	\end{axis}
\end{tikzpicture}
\caption{F-trigger, F-occupation}
    \label{fig:1}
  \end{subfigure}
  \begin{subfigure}{0.49\textwidth}
\begin{tikzpicture}
    \pgfplotsset{%
        width=\columnwidth,
        height=1.5in,
    }
    \tikzstyle{every node}=[font=\small]
	\begin{axis}[ybar stacked,bar width=4,
	symbolic x coords={\textbf{avg},de,pt,pl,cs,ru,lt,lv,uk,hr,fr,el,es,ro,hi,he,it,ur,ca,be,sr},
	axis x line*=bottom,
	axis y line*=left,
    xtick=data,
    ytick distance=0.2,
    ymin=0,
	nodes near coords align={vertical},
    x tick label style={rotate=90,anchor=east},
    ]
\addplot +[fill=green!40,draw=green!100] coordinates { (\textbf{avg},0.73) (de,0.83) (pt,0.81) (pl,0.83) (cs,0.82) (ru,0.85) (lt,0.87) (lv,0.79) (uk,0.78) (hr,0.73) (fr,0.80) (el,0.85) (es,0.68) (ro,0.80) (hi,0.79) (he,0.69) (it,0.62) (ur,0.73) (ca,0.67) (be,0.54) (sr,0.21) };

\addplot +[fill=yellow!40,draw=yellow!100] coordinates { (\textbf{avg},0.07) (de,0.01) (pt,0.02) (pl,0.02) (cs,0.02) (ru,0.01) (lt,0.01) (lv,0.01) (uk,0.02) (hr,0.03) (fr,0.07) (el,0.05) (es,0.17) (ro,0.04) (hi,0.08) (he,0.08) (it,0.23) (ur,0.20) (ca,0.19) (be,0.12) (sr,0.03) };

\addplot +[fill=red!40,draw=red!100] coordinates { (\textbf{avg},0.19) (de,0.16) (pt,0.17) (pl,0.15) (cs,0.16) (ru,0.14) (lt,0.11) (lv,0.20) (uk,0.20) (hr,0.24) (fr,0.14) (el,0.10) (es,0.15) (ro,0.16) (hi,0.13) (he,0.22) (it,0.15) (ur,0.07) (ca,0.14) (be,0.33) (sr,0.76) };

	\end{axis}
\end{tikzpicture}
\caption{M-trigger, F-occupation}
    \label{fig:1}
  \end{subfigure}
  \begin{subfigure}{0.49\textwidth}
\begin{tikzpicture}
    \pgfplotsset{%
        width=\columnwidth,
        height=1.5in,
    }
    \tikzstyle{every node}=[font=\small]
	\begin{axis}[ybar stacked,bar width=4,
	symbolic x coords={\textbf{avg},de,pt,pl,cs,ru,lt,lv,uk,hr,fr,el,es,ro,hi,he,it,ur,ca,be,sr},
	axis x line*=bottom,
	axis y line*=left,
    xtick=data,
    ytick distance=0.2,
    ymin=0,
	nodes near coords align={vertical},
    x tick label style={rotate=90,anchor=east},
    ]
\addplot +[fill=green!40,draw=green!100] coordinates { (\textbf{avg},0.28) (de,0.37) (pt,0.42) (pl,0.21) (cs,0.31) (ru,0.18) (lt,0.22) (lv,0.30) (uk,0.20) (hr,0.36) (fr,0.25) (el,0.19) (es,0.37) (ro,0.28) (hi,0.17) (he,0.22) (it,0.23) (ur,0.10) (ca,0.28) (be,0.32) (sr,0.68) };

\addplot +[fill=yellow!40,draw=yellow!100] coordinates { (\textbf{avg},0.10) (de,0.02) (pt,0.03) (pl,0.03) (cs,0.03) (ru,0.04) (lt,0.03) (lv,0.03) (uk,0.04) (hr,0.07) (fr,0.07) (el,0.10) (es,0.17) (ro,0.12) (hi,0.11) (he,0.16) (it,0.22) (ur,0.20) (ca,0.24) (be,0.21) (sr,0.09) };

\addplot +[fill=red!40,draw=red!100] coordinates { (\textbf{avg},0.62) (de,0.61) (pt,0.56) (pl,0.76) (cs,0.66) (ru,0.78) (lt,0.75) (lv,0.67) (uk,0.76) (hr,0.58) (fr,0.69) (el,0.72) (es,0.46) (ro,0.61) (hi,0.72) (he,0.63) (it,0.55) (ur,0.70) (ca,0.48) (be,0.47) (sr,0.23) };
	\end{axis}
\end{tikzpicture}
\caption{F-trigger, M-occupation}
    \label{fig:1}
  \end{subfigure}
\caption{Results for M2M model (1.2B). Proportion of correct (green), incorrect (red) and not available (yellow) are provided. Across the board, for all languages, gender inflection (green) are more correct for masculine triggers, MM (a) and MF (c) than feminine triggers FF (b) and FM (d). Accuracy is high for both masculine- and feminine- triggers when the the occupation is indicative of the target gender (a, b) than when it isn't (c,d). However, accuracy falls more more for F-triggers than for M-triggers when target occupation is indicative of the mismatched gender.}\label{fig:m2m1.2B}
  \end{figure*}
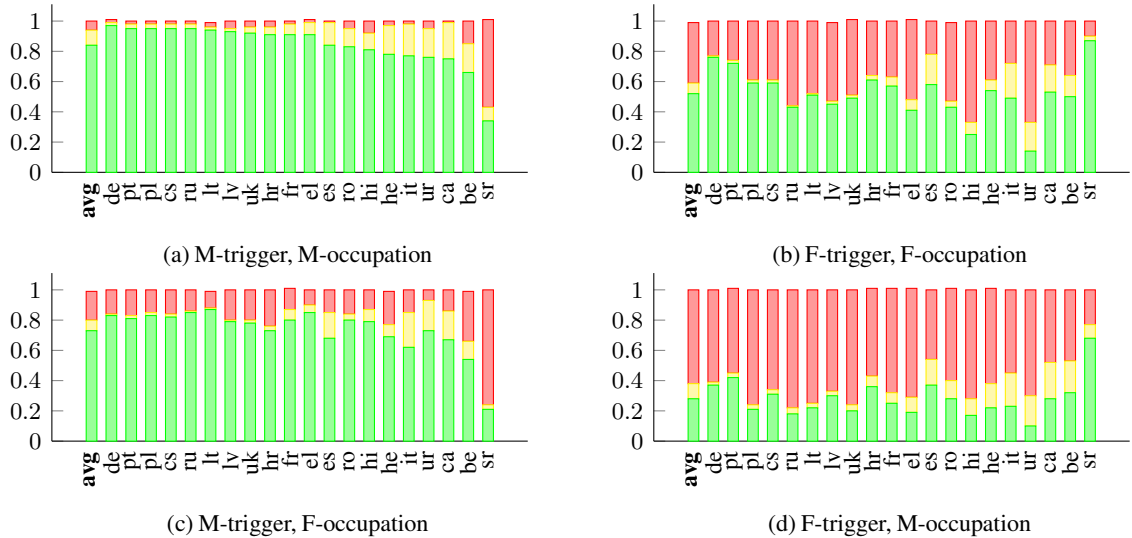
\captionsetup[]{font=small,skip=0pt}
\usetikzlibrary{patterns}
\begin{figure*}
\centering
  \begin{subfigure}{0.49\textwidth}
\begin{tikzpicture}
    \pgfplotsset{%
        width=\textwidth,
        height=1.5in,
    }
    \tikzstyle{every node}=[font=\small]
	\begin{axis}[ybar stacked,bar width=4,
	symbolic x coords={\textbf{avg},laborer,physician,engineer,sheriff,supervisor,lawyer,farmer,chief,carpenter,salesperson,developer,driver,plumber,manager,analyst,janitor,mover,guard,ceo},
	axis x line*=bottom,
	axis y line*=left,
    xtick=data,
    ytick distance=0.2,
    ymin=0,
	nodes near coords align={vertical},
    x tick label style={rotate=90,anchor=east},
    ]
\addplot +[fill=green!40,draw=green!90] coordinates { (\textbf{avg},0.84) (laborer,0.93) (physician,0.93) (engineer,0.92) (sheriff,0.92) (supervisor,0.91) (lawyer,0.90) (farmer,0.88) (chief,0.87) (carpenter,0.86) (salesperson,0.86) (developer,0.85) (driver,0.85) (plumber,0.83) (manager,0.82) (analyst,0.76) (janitor,0.76) (mover,0.74) (guard,0.72) (ceo,0.64) };

\addplot +[fill=yellow!40,draw=yellow!90] coordinates { (\textbf{avg},0.10) (laborer,0.02) (physician,0.04) (engineer,0.03) (sheriff,0.06) (supervisor,0.04) (lawyer,0.04) (farmer,0.06) (chief,0.11) (carpenter,0.06) (salesperson,0.05) (developer,0.07) (driver,0.14) (plumber,0.06) (manager,0.13) (analyst,0.17) (janitor,0.15) (mover,0.14) (guard,0.12) (ceo,0.33) };

\addplot +[fill=red!40,draw=red!90] coordinates { (\textbf{avg},0.06) (laborer,0.04) (physician,0.03) (engineer,0.05) (sheriff,0.02) (supervisor,0.05) (lawyer,0.05) (farmer,0.07) (chief,0.01) (carpenter,0.07) (salesperson,0.10) (developer,0.08) (driver,0.01) (plumber,0.11) (manager,0.05) (analyst,0.06) (janitor,0.09) (mover,0.12) (guard,0.17) (ceo,0.02) };

	\end{axis}
\end{tikzpicture}
\caption{M-trigger M-occupation}
    \label{fig:1}
  \end{subfigure}
  \begin{subfigure}{0.49\textwidth}
\begin{tikzpicture}
    \pgfplotsset{%
        width=\columnwidth,
        height=1.5in,
    }
    \tikzstyle{every node}=[font=\small]
	\begin{axis}[ybar stacked,bar width=4,
	symbolic x coords={\textbf{avg},housekeeper,nurse,secretary,librarian,teacher,receptionist,hairdresser,attendant,cleaner,baker,writer,cashier,assistant,accountant,counselor,clerk,editor,auditor,designer},
	axis x line*=bottom,
	axis y line*=left,
    xtick=data,
    ytick distance=0.2,
    ymin=0,
	nodes near coords align={vertical},
    x tick label style={rotate=90,anchor=east},
    ]
\addplot +[fill=green!40,draw=green!90] coordinates { (\textbf{avg},0.52) (housekeeper,0.89) (nurse,0.88) (secretary,0.71) (librarian,0.64) (teacher,0.64) (receptionist,0.62) (hairdresser,0.62) (attendant,0.55) (cleaner,0.54) (baker,0.51) (writer,0.48) (cashier,0.44) (assistant,0.43) (accountant,0.41) (counselor,0.40) (clerk,0.36) (editor,0.32) (auditor,0.29) (designer,0.21) };

\addplot +[fill=yellow!40,draw=yellow!90] coordinates { (\textbf{avg},0.07) (housekeeper,0.03) (nurse,0.07) (secretary,0.04) (librarian,0.02) (teacher,0.06) (receptionist,0.14) (hairdresser,0.05) (attendant,0.16) (cleaner,0.09) (baker,0.06) (writer,0.03) (cashier,0.04) (assistant,0.20) (accountant,0.11) (counselor,0.02) (clerk,0.14) (editor,0.03) (auditor,0.04) (designer,0.04) };

\addplot +[fill=red!40,draw=red!90] coordinates { (\textbf{avg},0.40) (housekeeper,0.09) (nurse,0.05) (secretary,0.24) (librarian,0.33) (teacher,0.30) (receptionist,0.23) (hairdresser,0.33) (attendant,0.29) (cleaner,0.37) (baker,0.42) (writer,0.49) (cashier,0.52) (assistant,0.37) (accountant,0.48) (counselor,0.58) (clerk,0.51) (editor,0.65) (auditor,0.67) (designer,0.76) };
	\end{axis}
\end{tikzpicture}
\caption{F-trigger F-occupation}
    \label{fig:1}
  \end{subfigure}
  \begin{subfigure}{0.49\textwidth}
\begin{tikzpicture}
    \pgfplotsset{%
        width=\columnwidth,
        height=1.5in,
    }
    \tikzstyle{every node}=[font=\small]
	\begin{axis}[ybar stacked,bar width=4,
	symbolic x coords={\textbf{avg},counselor,writer,editor,designer,cashier,auditor,accountant,baker,teacher,clerk,cleaner,hairdresser,librarian,assistant,attendant,secretary,receptionist,housekeeper,nurse},
	axis x line*=bottom,
	axis y line*=left,
    xtick=data,
    ytick distance=0.2,
    ymin=0,
	nodes near coords align={vertical},
    x tick label style={rotate=90,anchor=east},
    ]
\addplot +[fill=green!40,draw=green!90] coordinates { (\textbf{avg},0.74) (counselor,0.92) (writer,0.91) (editor,0.90) (designer,0.90) (cashier,0.86) (auditor,0.86) (accountant,0.83) (baker,0.83) (teacher,0.81) (clerk,0.80) (cleaner,0.77) (hairdresser,0.76) (librarian,0.75) (assistant,0.71) (attendant,0.67) (secretary,0.62) (receptionist,0.61) (housekeeper,0.37) (nurse,0.06) };

\addplot +[fill=yellow!40,draw=yellow!90] coordinates { (\textbf{avg},0.07) (counselor,0.01) (writer,0.04) (editor,0.03) (designer,0.03) (cashier,0.04) (auditor,0.03) (accountant,0.09) (baker,0.06) (teacher,0.05) (clerk,0.13) (cleaner,0.10) (hairdresser,0.04) (librarian,0.02) (assistant,0.19) (attendant,0.17) (secretary,0.04) (receptionist,0.16) (housekeeper,0.02) (nurse,0.07) };

\addplot +[fill=red!40,draw=red!90] coordinates { (\textbf{avg},0.19) (counselor,0.06) (writer,0.05) (editor,0.07) (designer,0.06) (cashier,0.10) (auditor,0.11) (accountant,0.08) (baker,0.10) (teacher,0.14) (clerk,0.07) (cleaner,0.13) (hairdresser,0.20) (librarian,0.23) (assistant,0.10) (attendant,0.15) (secretary,0.35) (receptionist,0.23) (housekeeper,0.60) (nurse,0.87) };
	\end{axis}
\end{tikzpicture}
\caption{M-trigger F-occupation}
    \label{fig:1}
  \end{subfigure}
  \begin{subfigure}{0.49\textwidth}
\begin{tikzpicture}
    \pgfplotsset{%
        width=\columnwidth,
        height=1.5in,
    }
    \tikzstyle{every node}=[font=\small]
	\begin{axis}[ybar stacked,bar width=4,
	symbolic x coords={\textbf{avg},laborer,guard,salesperson,plumber,carpenter,janitor,lawyer,mover,farmer,supervisor,developer,analyst,engineer,physician,manager,ceo,chief,sheriff,driver},
	axis x line*=bottom,
	axis y line*=left,
    xtick=data,
    ytick distance=0.2,
    ymin=0,
	nodes near coords align={vertical},
    x tick label style={rotate=90,anchor=east},
    ]
\addplot +[fill=green!40,draw=green!90] coordinates { (\textbf{avg},0.27) (laborer,0.53) (guard,0.47) (salesperson,0.46) (plumber,0.35) (carpenter,0.35) (janitor,0.32) (lawyer,0.31) (mover,0.30) (farmer,0.30) (supervisor,0.26) (developer,0.25) (analyst,0.25) (engineer,0.23) (physician,0.23) (manager,0.19) (ceo,0.14) (chief,0.09) (sheriff,0.09) (driver,0.08) };

\addplot +[fill=yellow!40,draw=yellow!90] coordinates { (\textbf{avg},0.10) (laborer,0.02) (guard,0.14) (salesperson,0.05) (plumber,0.07) (carpenter,0.07) (janitor,0.14) (lawyer,0.04) (mover,0.15) (farmer,0.06) (supervisor,0.05) (developer,0.08) (analyst,0.17) (engineer,0.03) (physician,0.04) (manager,0.13) (ceo,0.35) (chief,0.13) (sheriff,0.06) (driver,0.14) };

\addplot +[fill=red!40,draw=red!90] coordinates { (\textbf{avg},0.63) (laborer,0.44) (guard,0.39) (salesperson,0.48) (plumber,0.58) (carpenter,0.59) (janitor,0.54) (lawyer,0.66) (mover,0.55) (farmer,0.64) (supervisor,0.69) (developer,0.67) (analyst,0.58) (engineer,0.73) (physician,0.74) (manager,0.68) (ceo,0.51) (chief,0.78) (sheriff,0.85) (driver,0.78) };
	\end{axis}
\end{tikzpicture}
\caption{F-trigger, M-occupation}
    \label{fig:1}
  \end{subfigure}
\caption{Results for M2M model (1.2B). Proportion of correct (green), incorrect (red) and not available (yellow) are provided. Across the board, for all occupations, accuracy is higher when triggered gender matches the occupation (a, b), then when it mismatches (c, d). Additionally, accuracy is higher for masculine triggers (a, c) than for feminine ones (b, d).}\label{fig:m2m1.2B.occ}
  \end{figure*}
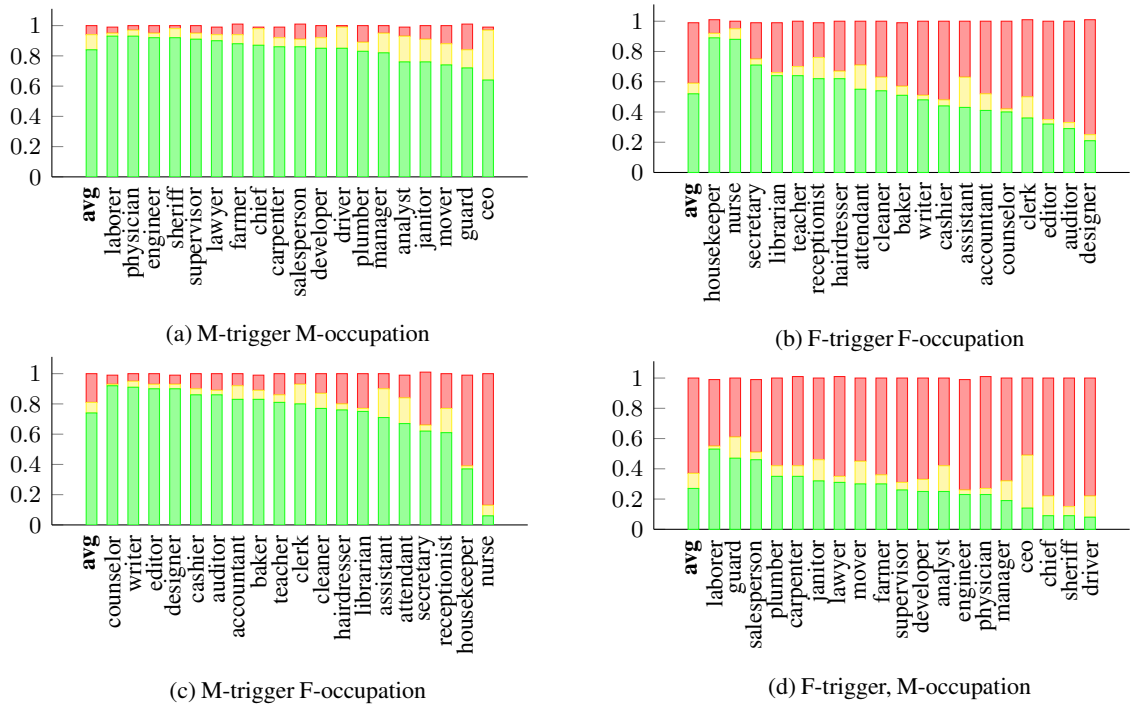


\section{Results}\label{sec:results}

\paragraph{Our dataset is very difficult for current transformer-based models.} We observe that accuracy doesn't exceed the low 70s for any language (see \autoref{tab:performance}). This suggests that our dataset is appreciably difficult, suggesting that it provides good signal about the failures of our current best models.

\begin{table}[t]
\small
\centering
\begin{tabular}{crrr}
\toprule
Language & \%Correct & \%Wrong &  \%N/A \\
\midrule
      de &     0.73 &   0.26 &  0.01 \\
      pt &     0.72 &   0.26 &  0.02 \\
      cs &     0.67 &   0.30 &  0.03 \\
      lt &     0.63 &   0.35 &  0.02 \\
      pl &     0.65 &   0.33 &  0.03 \\
      hr &     0.65 &   0.31 &  0.04 \\
      fr &     0.63 &   0.30 &  0.07 \\
      lv &     0.62 &   0.36 &  0.02 \\
      es &     0.61 &   0.21 &  0.17 \\
      ru &     0.60 &   0.38 &  0.02 \\
      uk &     0.60 &   0.37 &  0.03 \\
      el &     0.59 &   0.34 &  0.07 \\
      ro &     0.59 &   0.33 &  0.08 \\
      ca &     0.56 &   0.23 &  0.21 \\
      he &     0.55 &   0.32 &  0.13 \\
      it &     0.53 &   0.25 &  0.22 \\
      sr &     0.52 &   0.42 &  0.06 \\
      hi &     0.51 &   0.39 &  0.10 \\
      be &     0.51 &   0.33 &  0.17 \\
      ur &     0.44 &   0.37 &  0.19 \\
\bottomrule
\end{tabular}
    \caption{Aggregated accuracy for all languages and all data for M2M-100-1.2B.}
    \label{tab:performance}
\end{table}

\paragraph{For all languages and all MT systems, accuracy is higher when the trigger unambiguously refers to a man than when it unambiguously refers to a woman.}

In general, accuracy is lower when the trigger requires feminine morphology, hovering around 40\% in most languages. The only language for which accuracy on feminine triggers exceeds 50\%  is Serbian. For some languages, such as Urdu, occupation nouns are rarely inflected with the correct gender marking for feminine triggers. Taken in aggregate, these results likely reflect the cultural fact than many (but not all) languages utilize masculine to refer to generic people \citep{gastil1990generic, hamilton1991masculine}. Despite this, all our triggers are high frequency words, so we believe that a frequency based explanation of our findings won't be sufficient.

\paragraph{Accuracy is higher when trigger gender and occupation gender match, than when they don't\textellipsis} As we see in \autoref{fig:m2m1.2B}, the M2M models perform better on inflecting occupations nouns correctly when they are statistically more likely to refer to a person whose gender matches the gender required by the trigger: for example, our models is better at correctly morphologically gender marking \textit{nanny} (a statistically feminine-indicative occupation) in the context of \textit{mother} than they are at morphologically gender marking \textit{janitor} (a statistically masculine-indicative one). This finding replicates previous work \citep{stanovsky-etal-2019-evaluating} that showed that six then-state-of-the-art models were very susceptible to statistical gender biases encoded in occupation words. 

\paragraph{\textellipsis However, gender marking accuracy drops less when the occupation is mismatched with a masculine trigger than when it is mismatched with a feminine one.} Although statistical gender biases in how women are presented of the kind presented in \autoref{fig:m2m1.2B} are relatively well described in NLP and adjacent fields \citep{bolukbasi2016man,hovy2016, caliskan2017, rudinger-etal-2017-social,garg2018, garimella2019, gonen-goldberg-2019-lipstick, dinan-etal-2020-queens, dinan-etal-2020-multi}, we further observe in our results yet a higher order type of stereotyping that negatively affects women, namely \textit{androcentrism} \citep{bem1993lenses,hegarty2013androcentrism,  bailey2019man}. Androcentrism is a wide reaching cultural phenomenon that treats the ``male experience\textellipsis as a neutral standard or norm for the culture of the species as a whole'' \citep[p. 41]{bem1993lenses}---one consequence of this cultural phenomenon is that women are restricted to their stereotypical domains more than men are to theirs. We see some tentative evidence that our MT systems encode this cultural androcentrism bias in the fact that the drop in accuracy is greater for sentences with feminine triggers (e.g., \textit{mother}) and masculine-indicative occupations (\textit{janitor}) than for the converse (compare the magnitude of the drop in \autoref{fig:m2m1.2B} and \autoref{fig:m2m1.2B.occ} between a and c to the drop between b and d, as well as \autoref{tab:androcentrism}).

\paragraph{Models achieve higher accuracy for masculine-indicative than feminine-indicative occupation (and there is some variation).} Finally, to understand the behavior of particular occupations, we plot the M2M 1.2B accuracy by occupation averaged across all languages. Recall that all occupations are frequent, are either statistically biased towards either men or towards women in the source, and are balanced in the dataset. In \autoref{fig:m2m1.2B.occ}, we observe that in the case of feminine grammatical gender triggers, only a few feminine-indicative occupations (e.g. \textit{houskeeper, nurse, secretary} in \autoref{fig:m2m1.2B.occ} b, d) reach the level of accuracy that the model achieves on most masculine-indicative occupations (in \autoref{fig:m2m1.2B.occ} a, c). We also note that variation in accuracy is much higher for feminine-indicative occupations across both trigger types (compare \autoref{fig:m2m1.2B.occ} b to c). These results also lend support to a cultural androcentrism hypothesis. 


\section{Related Work}\label{sec:relatedwork}

Recently, several works \citep{stanovsky-etal-2019-evaluating,prates2019assessing, gonen-webster-2020-automatically, gonzalez-etal-2020-type} investigated gender bias in multiple languages with complex morphology, and showed that state-of-the-art MT systems resolve gender-unbalanced occupation nouns (from the US Bureau of Labor Statistics) more often to masculine than feminine pronouns, despite the fact that people of many genders participate in all listed occupations. Our work improves upon these prior approaches by exploring the effects of gender-indicative contexts (e.g., additionally stereotypically masculine and feminine traits and events) in range of syntactic positions (e.g., preceding or following the clue, directly adjacent to the occupation, etc.). While \newcite{prates2019assessing} did investigate some stereotypical traits in their work, they only investigate a few of them, only in the  context of the ambiguous paradigm, and were narrowly focused on measuring the translation abilities of one commercial translation product. We, on the other hand, explore not only more diverse example traits as well as additional verbal contextual cues, but we do so in unambiguously gendered sentences with a diverse range of sentences structures that allow us to vary linear precedence of contextual cues as well as their prevalence. \newcite{gonen-webster-2020-automatically} also made use of minimally different sentences via an innovative perturbation method that mines examples from real world data and moves away from static word lists; however, their benchmark is also collected for the standard ambiguous gender setting.  

\begin{table}[h]
\small
\centering
\begin{adjustbox}{max width=\columnwidth}
\begin{tabular}{crrrr}
\toprule
& \multicolumn{2}{c}{M2M-100-1.2B} & \multicolumn{2}{c}{Opus MT}\\
Language &  $\Delta M$ &  $\Delta F$ &  $\Delta M$ &  $\Delta F$ \\
\midrule
      be &     0.17 &  \bf  0.25 \\
      ca &     0.13 &  \bf  0.25 &      0.20 &  \bf  0.29 \\
      cs &     0.23 &  \bf  0.33 &      0.16 &  \bf  0.19 \\
      de &     0.16 &  \bf  0.30 &      0.16 &  \bf  0.22 \\
      el &     0.08 &  \bf  0.19 \\
      es &     0.14 &  \bf  0.27 &      0.14 &  \bf  0.20 \\
      fr &     0.15 &  \bf  0.25 &      0.16 &  \bf  0.24 \\
      he &     0.00 &  \bf  0.29 &     -0.02 &  \bf  0.29 \\
      hi &    -0.04 &    0.02 \\
      hr &     0.22 &  \bf  0.28\\
      it &     0.11 &  \bf  0.23 &      0.17 &  \bf  0.23 \\
      lt &     0.05 &  \bf  0.10\\
      lv &     0.15 &  \bf  0.20 \\
      pl &     0.19 &  \bf  0.34 \\
      pt &     0.15 &  \bf  0.25\\
      ro &     0.12 &  \bf  0.17 &      0.14 & \bf   0.22 \\
      sr &     0.16 &  \bf  0.23 \\
      uk &     0.19 &  \bf  0.30 &      0.12 &  \bf  0.19 \\
      ur &    -0.02 &   -0.01 & & \\ 
      
\bottomrule
\end{tabular}
    \end{adjustbox}
    \caption{Accuracy drop ($\Delta$) is greater for feminine triggers when the occupation is statistically indicative of the mismatched gender (M) than for masculine triggers when the occupations is statistically indicative of the mismatched gender (F) For example, $\Delta M$ refers to the drop in accuracy for sentences with triggers referring to men when the occupation is switched to stereotypically referring to women (e.g., difference in accuracy between ``my father is a funny doctor'' and ``my father is a funny nurse''. Bold marks the delta with larger accuracy drop. }
    \label{tab:androcentrism}
\end{table}

Of particular note here is \newcite{gonzalez-etal-2020-type} which also focused on ``unforgivable'' grammatical gender-related errors in translation (as well as on other tasks) that come about as a result of syntactic structure and unambiguous coreference. In particular, \citeauthor{gonzalez-etal-2020-type} investigated four languages (Danish, Russian, Chinese, Swedish) that affix a reflexive marker to disambiguate whether a 3rd person possessive pronoun (e.g., \textit{his}) must be obligatorily bound by its local referent (i.e., the subject of the sentence) or not. This approach is somewhat analogous to some of our examples, except that we rely on syntactic context to construct unambiguous examples as opposed to language-internal properties: e.g., particularly those that make use of ``own'' to make obligatory the local coreference (in this case cataphora) as in ``That her \textit{own} child cried, surprised the doctor''. We take our work to be wholly complementary to theirs; while their approach focuses on more source languages, fewer target languages, and a wider range of tasks, we focus on fewer source languages, more target languages, and sentences from a wider range of (source) syntactic structures (as determined by our grammar).


Concurrently, another approach to pronoun coreference utilized a hand-crafted grammar to generate sentences for measuring fairness \citep{soremekun2020astraea}, but in the context of NLP tasks other than NMT. Although \newcite{soremekun2020astraea} are interested in measuring performance for unambiguous examples, it does not focused on the NMT use case, and its examples require cross-sentential coreferences, which will likely require a more complex linguistic toolbox than our intrasentential case \citep{szabolcsi2003binding, hardmeier2010modelling, reinhart2016anaphora}. Moreover, the grammar created in that work is much less developed than ours: it does not manipulate the location of the trigger, there is limited syntactic diversity, and there is no incorporation of statistically gender-biased words above and beyond occupation nouns.

At a high level, our work resurfaces problems with syntactic agreement in machine translation. While neural machine translations are more fluent that phrase-based machine translation, it has long been observed that even high-resource models can struggle to generate faithful translations that are also syntactically correct~\cite{isabelle-etal-2017-challenge} and the problem intensifies for longer sentences with long-distance dependencies~\cite{choshen-abend-2019-automatically}. We highlight yet another syntactic failure mode in NMT models in this work.
There is also a long history of incorporating syntax explicitly into NMT models in the hope of reducing the prevalence of such errors. For example, \newcite{eriguchi-etal-2016-tree} model source-side syntax while \newcite{aharoni-goldberg-2017-towards} proposed models that generate linearized dependency trees. Other works also consider modifications to the attention mechanism in order to improve NMT~\cite{kim2017structured}.
\Adi{added a few syntax papers, there are many many more, if we need I can add some more...}

\section{Conclusion}

Many of our NLP tasks and dataset have been found to be rife with statistical gender biases that reflect, in language, the stereotypical associations we have about gender in our cultures. In this work, we present a new evaluation dataset for measuring gender bias in machine translation for gender unambiguous sentences. Our dataset supports translation from an English source into \howmanylanguages\ languages, and is designed to answer questions not only about particular occupation words and gender triggering words, but also to further explicate the role of context in how MT systems translate gender morphology. We hope that our dataset will encourage the community to improve on this new setting for measuring gender biases in language.


\appendix

\begin{table}[t]
\small
\begin{tabular}{p{5em}p{9em}p{9em}}
\toprule
Type &  F & M   \\
\midrule
  Context\textsubscript{Adj} & affected, affectionate, appreciative, emotional, excitable, imaginative, impressionable, intelligent, organized, outgoing, unambitious, understanding, unintelligent, unselfish, unstable, cautious, changeable, charming, cheerful, childlike, clean, compassionate, complaining, complicated, confused, cooperative, creative, critical, curious, dainty, delicate, dependent, dreamy, family-oriented, fashionable, fault-finding, fearful, feminine, fickle, flatterable, flirtatious, foolish, forgiving, friendly, frivolous, fussy, gentle, graceful, gullible, helpful, honest, kind, loyal, melodramatic, mild, modest, naive, nervous, patient, pleasant, polite, prudish, romantic, self-pitying, sensitive, sentimental, sexy, short, shy, small-boned, smart, soft, softhearted, sophisticated, spiritual, submissive, suggestive, superstitious, sympathetic, talkative, tender, timid, touchy, warm, weak, well-dressed, well-mannered, wholesome, worrying, yielding
 & aggressive, active, adventurous, aggressive, ambitious, analytical, arrogant, assertive, athletic, autocratic, enterprising, independent, indifferent, individualistic, initiative, innovative, intense, inventive, obnoxious, opinionated, opportunistic, unfriendly, unscrupulous, bossy, broad-shouldered, capable, coarse, competitive, conceited, confident, consistent, controlling, courageous, cruel, cynical, decisive, demanding, dependable, determined, disciplined, disorderly, dominant, forceful, greedy, hardhearted, hardworking, humorous, jealous, lazy, level-headed, logical, loud, masculine, muscular, pleasure-seeking, possessive, precise, progressive, promiscuous, proud, quick, rational, realistic, rebellious, reckless, resourceful, rigid, robust, self-confident, self-reliant, self-righteous, self-sufficient, selfish, serious, sharp-witted, show-off, solemn, solid, steady, stern, stingy, stolid, strong, stubborn, sturdy, tall, tough, well-built, witty \\
 Context\textsubscript{V-OBJ} & protect, treat, shame, exploit insult, scare, frighten, distract, escort & reward, glorify, thank, praise, honor, inspire, enrich, appease, congratulate, respect, flatter, destroy, deceive, bore, offend, scold, pay, fight, defeat \\
 Context\textsubscript{V-SUBJ} & smile, dance, laugh, play, giggle, weep, faint, scream, gossip, complain, lament, spin, celebrate, clap & succeed, flourish, prosper, win, protest, kill, threaten, rush, speak\\
\bottomrule
\end{tabular}
    \caption{Gendered context words from our dataset.}
    \label{tab:contextcues}
\end{table}

\Adi{
\section{Things to do}
\begin{itemize}
    \item get results from opus-mt, m2m-small, and m2m-medium (currently) (adi, by tomorrow)
    \item ping angela for m2m-large (currently) (Adi, by monday)
    \item get occupation translations from Luke's paper/data. (Adi)
    \item talk to team PoCs for annotations regarding budget. (Adi)
    \item Look up aws for crowdsourcing (Adi)
    \item script for evaluation, we have to decide what the evaluation entails (Adi, Adina)
    \item Find a small pretrained model to run. (Adi)
    \item expert annotation of Luke's templates for ambiguity and our templates for ambiguity (Adina)
    \begin{itemize}
        \item Optional: make templates for questions
    \end{itemize}
\end{itemize}

\noindent Timeline:
\begin{itemize}
    \item internal review deadline: Apr 7
    \begin{itemize}
        \item eval scripts done by Apr 2 EOD
        \item 3 MT models (3 divisions in generated/s folder)
    \end{itemize}
    \item arxiv Apr 17
        \begin{itemize}
            \item Angela's help on finetuning?
            \item dict check for German/Spanish
            \item model size comparison
        \end{itemize}
    \item emnlp May 17
        \begin{itemize}
            \item other source langs? (no pronoun/rel noun clues), perhaps french, perhaps german
            \item consider expanding dict. check
            \item get xl working
        \end{itemize}
\end{itemize}

Tables: (anti-only, acc per occ., avg acc per occ)
\begin{itemize}
    \item Table 1: explicit CLUE token (which gender is explicitly provided e.g., by mother/her),  occupation (whether it is man/woman indicative) : 4 columns
    \item Table 2: does CLUE precede/follow occupation noun?
    \item Table 3: number of suggestive CLUES (adj/verb: 0 none, 1 adj, 1 verb, 2 adj+verb).
    \item Table 4: observation that smaller model and larger model do comparably(?) on fewer languages
    \item Table 5: correctness of gender marking on adj and verb for languages who have those
    \item Table 6 (placeholder): what happens with unseen occupations (might need to do training here)
\end{itemize}

Models:
\begin{itemize}
    \item Many-2-Many (SOTA) - rope in Angela to help with finetuning? if no, train our own (w/OPUS) - finetune on the dataset?
    \item small pretrained model?
    \item ``system a" microsoft, ``system b" google (unofficial package)
    \item train our own (if we want to answer Tab 6 question)
\end{itemize}
}

\bibliography{anthology,emnlp2020}
\bibliographystyle{acl_natbib}
\end{document}